\documentclass[journal,letterpaper]{IEEEtran}
\usepackage[T1]{fontenc}
\usepackage{amsmath,amssymb,graphicx,booktabs,array}
\usepackage{cite}
\usepackage[hidelinks]{hyperref}
\usepackage{url}
\hypersetup{pdftitle={Reusing Scene-Graph Readouts Across Asynchronous Robot Updates},
  pdfauthor={Yilun Wu, Ying Liu, Yunjian Zhang, Aobo Li, Mu-Jiang-Shan Wang}}
\begin{document}
\bstctlcite{IEEEexample:BSTcontrol}
\title{Reusing Scene-Graph Readouts Across Asynchronous Robot Updates}
\author{Yilun~Wu, Ying~Liu, Yunjian~Zhang, Aobo~Li, and Mu-Jiang-Shan~Wang,~\IEEEmembership{Member,~IEEE}%
\thanks{\raggedright Yilun Wu, Yunjian Zhang, and Aobo Li are with Shenzhen Kaihong Digital Industry Development Co., Ltd., Shenzhen, China. Y. Wu (\href{mailto:wuyilun310@gmail.com}{\mbox{wuyilun310@gmail.com}}, ORCID \href{https://orcid.org/0000-0001-5169-4864}{\mbox{0000-0001-5169-4864}}). Y. Zhang (\href{mailto:yunjzhang@xmu.edu.cn}{\mbox{yunjzhang@xmu.edu.cn}}, ORCID \href{https://orcid.org/0000-0002-0303-3671}{\mbox{0000-0002-0303-3671}}). A. Li (\href{mailto:ali09@qub.ac.uk}{\mbox{ali09@qub.ac.uk}}, ORCID \href{https://orcid.org/0009-0001-6868-9495}{\mbox{0009-0001-6868-9495}}).}%
\thanks{\raggedright Ying Liu is with Southern University of Science and Technology, Shenzhen, China (\href{mailto:Liuy63@sustech.edu.cn}{\mbox{Liuy63@sustech.edu.cn}}, ORCID \href{https://orcid.org/0000-0001-6016-3184}{\mbox{0000-0001-6016-3184}}).}%
\thanks{\raggedright Mu-Jiang-Shan Wang (corresponding author) is with Shenzhen Kaihong Digital Industry Development Co., Ltd., Shenzhen, China, and with Harbin Institute of Technology, China (\href{mailto:wangmujiangshan-qy@kaihong.com}{\mbox{wangmujiangshan-qy@kaihong.com}} and \href{mailto:wmjsdr0@gmail.com}{\mbox{wmjsdr0@gmail.com}}, ORCID \href{https://orcid.org/0000-0003-0950-4558}{\mbox{0000-0003-0950-4558}}).}}
\maketitle
\begin{abstract}
A robot policy may finish decoding after its input scene graph has changed. We test whether a delayed readout still equals the readout on the latest committed snapshot. The runtime propagates local edits through a message-passing encoder and stamps the final-layer rows read by each request. For fixed local layers and declared inputs, a structural induction gives a sufficient condition for exact reuse. On a 4,097-node spatial lattice driven by recorded gripper traces, the median update time fell from 3.81 ms for full recomputation to 0.097 ms for incremental refresh. Six manipulation tasks with only two to eight tracked objects gave the opposite result. Including thread handoff, the learned motion head and software receipt, the 99th percentile was 0.309 ms for preallocated full recomputation and 0.446 ms for directed local refresh. Under two directed update traces with reordered inputs, slot replacements and model changes, the guard released 221 of 654 delayed readouts. Every release agreed with an independent full oracle. In a separate 64-slot stress test, checking object identity, model and decoder context together with local edits released 5,053 of 15,984 delayed requests without a stale acceptance. For small prebuilt host graphs, a captured GPU forward pass including transfers was slower than the CPU and drew 25.41 W at the discrete board under sustained execution. Shared-slot arbitration also required comparable clocks across sensors.
\end{abstract}

\section{Introduction}
\IEEEPARstart{A}{ robot} sends a graph embedding to a decoder that is still working when the next observation arrives. Suppose the new observation moves a cup at the far end of a mapped room. If the decoder read only the local tray, its original input may be exactly the same on the new snapshot. If the observation moved the tray itself, the decision can depend on stale evidence. The runtime needs to distinguish these cases without executing the decoder twice.

PI-VLA uses action consensus and prediction error to decide when to replan \cite{pi}. Real-time action chunking \cite{realtime}, masked action chunking \cite{remac} and streaming VLA decoding \cite{flash} all overlap inference and control in different ways. They expose a systems question upstream of action quality: at release time, is the graph readout on which this request depends still the value a fresh encoder would return? An unchanged readout says nothing about unobserved changes in the world. It does, however, permit exact reuse within a declared graph interface.

We provide an executable implementation of that interface and separate its two costs. One cost is refreshing the node embeddings affected by an update. The other is determining which delayed reads are still valid. Incremental graph inference already has efficient algorithms \cite{stream,ripple,d3}. Our contribution is the connection between per-layer invalidation and release-time validation, together with a replay study that measures when maintaining this information is worthwhile. Exact local refresh is prior art. Our reference implementation is deliberately small enough to compare after every update against a complete graph pass.

We use two types of recorded motion. Public object tracks provide naturally small graphs, including stationary periods. A separate workload uses public end-effector traces to move a hand node across a controlled map of up to 4,096 cells. That map is generated by the experiment. Its topology, read requests and decoder delays are known, so a stale acceptance can be detected directly. A feature-burst experiment tests how quickly spatial selectivity disappears as simultaneous edits increase.

\section{Relation to Existing Systems}
\subsection{Producing and consuming a robot graph}
Dynamic scene graphs expose objects, places and agents to a robot \cite{sg}. Kimera and Hydra connect spatial perception to continuously maintained scene structure \cite{kime,hydra}. ConceptGraphs represents open-vocabulary objects and relations \cite{concept}. MomaGraph associates spatial and functional structure with the current task \cite{moma}; Robot-WATCHDOG uses a graph of tracked objects to detect anomalous behavior \cite{watch}. Each supplies a reason to update a graph while an inference request is outstanding. The meaning of an update depends on which signals the upstream system publishes.

PI-VLA and OpenVLA \cite{pi,openvla} consume sensor information at policy rate. Octo and RT-2 illustrate how broad an input interface can become \cite{octo,rt2}. BagelVLA alternates planning, forecasting and action \cite{bagel}, while Motus unifies action and world modeling \cite{motus}. A global visual token updated at every frame can invalidate a request even when its object nodes stay fixed. A practical integration therefore needs an explicit declaration of every input to the decoder. The local guarantee is meaningful only when the consumer honors that declaration. Model expansion under uncertainty \cite{hypo} provides another example where changing state structure has consequences for downstream planning.

\subsection{Incrementality, ordering, and locality}
InkStream exploits the incremental structure of graph aggregation \cite{stream}; Ripple supports exact incremental inference for large streaming graphs \cite{ripple}; D3-GNN processes dynamic graphs in a distributed dataflow system \cite{d3}. Earlier systems such as Naiad expose logical time to incremental computations \cite{naiad}. The present CPU prototype recomputes whole affected rows in canonical order. It is a reference for freshness experiments rather than a replacement for the optimized kernels in these systems. A direct speed comparison to those implementations would require matched operators, update streams and hardware.

Logical clocks and consistent snapshots define how observations are ordered \cite{ddia,snapshot}. Here all graph writes pass through one sequencer. That simplification avoids distributed-clock reconstruction. In robot perception, iSAM2 exploits localized updates to a factorization \cite{isam}, factor graphs encode dependencies explicitly \cite{gtsam}, and switchable constraints allow individual pose-graph observations to lose influence \cite{switch}. A changed pose estimate can nevertheless touch a very large number of downstream graph rows. Locality is a property of both the update and the readout.

Message-passing layers have finite neighborhoods \cite{mpnn,inter,sage}. A rotating set of learned weights, as in EvolveGCN \cite{dyn}, or global attention \cite{rgraph} adds dependencies outside that neighborhood. Graph acceleration also depends on memory traffic and kernel scheduling \cite{pica}. OASIS addresses placement and scheduling across device, edge and cloud \cite{oasis}; our measurements hold placement fixed. Thermal feedback control \cite{thermal} is an example of an application in which a stale sensor value and a slow prediction have distinct operational meanings.

Several failure modes occur before the encoder runs. A changed bipartite association \cite{hybrid} can replace an object's identity without changing the number of visible objects; a multimodal detector can change node attributes \cite{mfe}. Work on interconnection-network diagnosis distinguishes the fault set from the observations available to distinguish it \cite{diag,good,leaf}. Our stamps identify which declared computations might have changed. They cannot diagnose an unreported sensor defect. Evidence-chain propagation in HyCE-RAG \cite{hyce} likewise makes upstream dependencies explicit, although its application is document retrieval. These distinctions determine what a release certificate can legitimately cover.

\section{Snapshot Contract}
Let $G_v=(X_v,A_v)$ be a committed graph with monotonically increasing integer version $v$. A row of $A_v$ lists the incoming neighbors of one receiver. A fixed $K$-layer encoder computes
\begin{equation}
 H^0_v=X_v,\qquad
 H^\ell_v[i]=F_\ell\!\left(H^{\ell-1}_v[i],
             \{H^{\ell-1}_v[j]:j\in A_v[i]\}\right).
 \label{eq:enc}
\end{equation}
Any edge attributes, normalization terms or masks read by $F_\ell$ must be part of the declared input. The code implements a mean of neighbor features, deterministic weights, a self term and a $\tanh$ activation. Degrees are receiver-local. A symmetric degree normalization that also reads the sender's degree would need to invalidate additional receivers when that degree changes.

At issue time $s$ a request records the IDs $R$ of all final-layer rows it reads, and the associated graph version. The tested readout is one row, which avoids any hidden dependence on an order of variable-length arrays. A deterministic decoder reading only those rows and fixed auxiliary inputs produces the same output when all those rows have the same values. A randomized decoder also needs its randomness held fixed for equality of its output. If it samples separately, equality of the inputs yields equality of conditional distributions.

\subsection{A receiver-oriented update}
Write $C_0$ for rows whose input features changed, and $B$ for receivers whose incoming edge lists or local edge attributes changed. Let $N_v^+(C)$ denote receivers with a predecessor in $C$ under the \emph{new} adjacency. Define
\begin{equation}
 C_\ell=B\cup C_{\ell-1}\cup N_v^+(C_{\ell-1}),
 \quad 1\le\ell\le K .
 \label{eq:dirty}
\end{equation}
For each layer, recompute the rows in $C_\ell$ against the updated previous layer. Keep unaffected rows from the earlier snapshot, then write the commit version to the final-layer stamps for nodes in $C_K$. All reads used by a request must see one committed version. Our serial implementation provides this atomicity directly. A parallel implementation would require a barrier, immutable buffers or an equivalent commit protocol.

The term $B$ must appear at every layer. A deleted edge can change a receiver's aggregation even if the sender's current input did not change. Propagating only from nodes with new features misses that receiver. Enumerating only present neighbors of the moved hand can also omit the old attachment cell. The replay captures both the previous and new attachment in $B$ before processing the layers.

\begin{figure*}[!t]\centering\includegraphics[width=.84\textwidth]{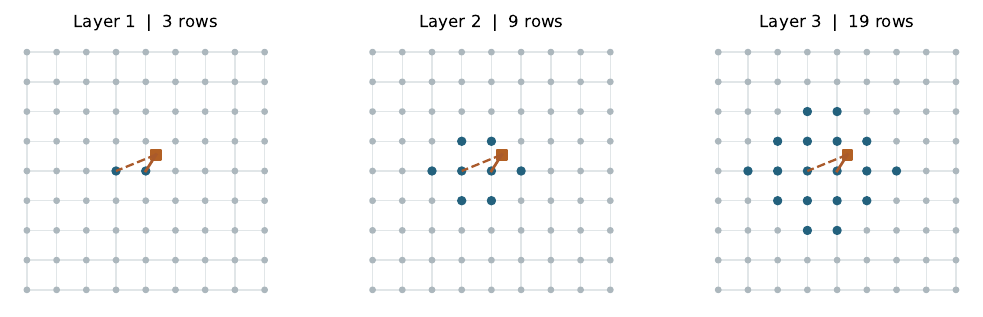}
\caption{Rows scheduled for recomputation in a nine-by-nine example after one hand attachment moves. The square is the hand node. The dashed and solid edges show its old and new attachments. Recomputed rows (blue) expand one hop per layer. The uncolored nodes remain available to outstanding local reads.}
\label{fig:cone}\end{figure*}

\subsection{Release-time proof}
\emph{Proposition 1.} For a deterministic encoder satisfying \eqref{eq:enc}, the above update yields the same mathematical embeddings as full recomputation. If each read node $i\in R$ has final stamp $\sigma_i\le s$, its current embedding equals its issue-time value.
\begin{IEEEproof}
At layer zero, all changed rows are in $C_0$. A receiver outside $C_\ell$ has the same incoming list, the same previous self row and unchanged previous rows for all current neighbors. Its inputs to $F_\ell$ are identical. Rows inside $C_\ell$ are explicitly recomputed from the correct previous layer. Induction proves equality with a fresh snapshot. For the second statement, any row possibly changed by any commit after $s$ would have received a stamp greater than $s$. A row passing the stamp check was outside every such affected set. Applying the first part at each commit proves its value was preserved.
\end{IEEEproof}

Stamp equality is sufficient. Canceling changes, saturation or zero weights may preserve a value even when a row was scheduled. The method may reject those requests. It also avoids a subtle ordering error: checking whether the latest \emph{global} version was touched during decoding is too coarse, while checking only the most recent edit misses an earlier invalidation. One monotone per-row stamp retains the necessary history. Concurrent release and update would need a defined linearization point. The guarantee is about that snapshot, not all subsequent robot states.

For row width $d$ and maximum fan-in $\Delta$, recomputation of one affected row costs $O(d^2+\Delta d)$ for the operator used here. Ignoring set-management overhead, the total is proportional to $\sum_\ell |C_\ell|(d^2+\Delta d)$. The implementation also sorts and deduplicates candidate IDs. It caches $(K+1)(n+1)d$ floating-point values including a padded zero row. Checking an outstanding request costs $O(|R|)$ integer comparisons. A change to global model weights or map coordinates can eliminate all reuse.

\section{Implementation and Workloads}
\subsection{Measurement boundaries}
The prototype uses Python 3.12.15, NumPy 2.5.5 and float64 on a Windows 11 host with an AMD Ryzen 9 8940HX CPU. Both paths limit BLAS to one thread. They use identical weight matrices, a fixed $\tanh$ operator, three layers and width 12. We draw fixed weights from a zero-mean Gaussian with standard deviation 0.18. Freshness and arithmetic cost can be evaluated without task training; the readouts have no calibrated behavioral meaning. Consistent ordering of padded neighbors keeps the summations comparable.

We timed full forward computation and incremental cache maintenance alternately first on successive versions with a high-resolution monotonic clock. The incremental timer includes affected-set construction and stamp writes. The full timer includes allocating new outputs. The same input decoding, graph editing, logging and differential check are outside both timers. We dropped the first four updates per episode from timing summaries and retained them for correctness checks. We report medians and 95th percentiles across individual updates, not hard deadlines. Single-machine measurements are sensitive to allocation strategy, vectorization and background scheduling.

\subsection{Observed object tracks}
We took the first ten nominal episodes in file order for each of six Robot-WATCHDOG tasks \cite{watch}. These cover conveyor sorting, table setting and household sorting or tidying. We read the public centroid tracks and linked distinct objects within 80 image pixels. Two feature coordinates contain position divided by 512, and the remaining coordinates are fixed episode-specific features. Edges are undirected and represented by two incoming entries. Each episode contributes up to 64 approximately evenly spaced updates over the available common track length. All resulting graphs contained two, four, six or eight objects. This workload uses measured image trajectories and a constructed proximity rule. No failure labels enter the experiment.

\begin{figure*}[!t]\centering\includegraphics[width=.93\textwidth]{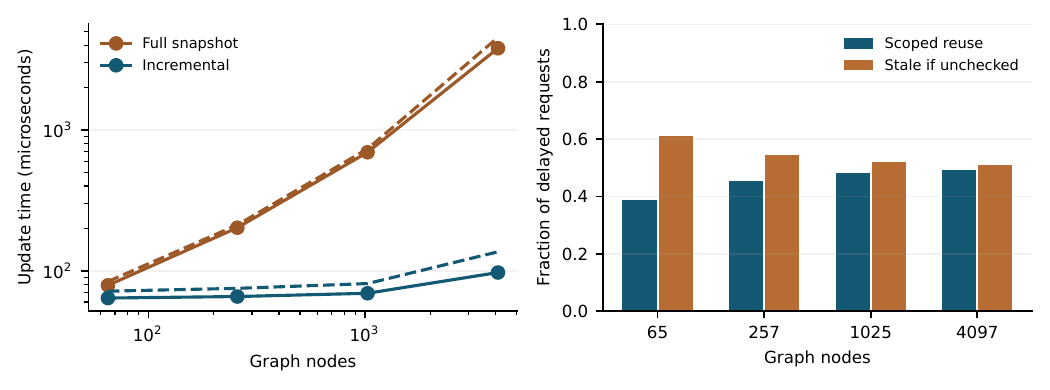}
\caption{Map replay. Solid lines show median update time and dashed lines the 95th percentile. On the right, reuse and stale fractions are computed on all completed delayed requests; their sum can be less than one when the structural rule rejects a numerically unchanged value.}
\label{fig:perf}\end{figure*}

\subsection{Recorded gripper motion on a generated map}
We used a pinned LeRobot conversion of LIBERO-10 demonstrations \cite{libero}. For each of the ten tasks we selected the two lowest episode IDs with at least 64 even-numbered frames among frames 0--127. The hand node receives the observed eight-component state, three squared position coordinates and a constant. A square grid covers $[-0.45,0.45]^2$ in the first two position dimensions. Coordinates are clipped to the map and quantized to a cell. Grid cells have four-neighbor connectivity. One undirected edge attaches the hand to its current cell. Other cell features are fixed random values for that episode. The source data supply the hand motion. They do not supply the spatial lattice or labels for its cells.

Map widths were 8, 16, 32 and 64 cells. Every transition creates a request on the pre-update snapshot. Odd requests read the current attachment and even requests read a uniformly chosen cell. Each request is validated after three updates, including the first update immediately after issue. The last two requests of an episode have insufficient future versions and are omitted from the release analysis. We compare the stored embedding with the exact fresh one at release, calling a difference stale if its largest coordinate exceeds $10^{-10}$. A global-version policy rejects every request in this workload. A no-validation policy releases all. A third, retrospective oracle can compare the actual embedding values after full recomputation; it provides an upper bound on reuse at the numerical tolerance.

\section{Results}
\begin{table}[!t]\centering\caption{LIBERO-driven map replay. Times are microseconds over 1,180 timed updates per size. Work is the fraction of layer rows refreshed.}\label{tab:tim}
\footnotesize\begin{tabular}{@{}rrrrrr@{}}\toprule Nodes&Full p50&Full p95&Local p50&Local p95&Work\\\midrule
65&78.9&84.0&64.1&71.7&11.66\%\\
257&202.4&211.9&65.7&75.1&3.07\%\\
1,025&695.1&734.9&69.3&81.2&0.82\%\\
4,097&3,808.4&4,550.4&97.3&136.5&0.22\%\\\bottomrule\end{tabular}\end{table}

All 5,040 lattice transitions produced the same final embeddings as full recomputation at the resolution of the stored float64 arrays. Local work varied little with map size; each version edited a hand feature and at most two cell attachments. Full forward time grew with the number of cells. The ratio of medians ranged from 1.23 at 65 nodes to 39.14 at 4,097. That last ratio includes the reference implementation's full-pass allocation cost. It should not be interpreted as an end-to-end policy acceleration or a speedup over InkStream or Ripple.

\begin{table}[!t]\centering\caption{Delayed reads out of 1,220 per map. The value oracle counts reads equal at the stated tolerance.}\label{tab:reads}
\footnotesize\begin{tabular}{@{}rrrrr@{}}\toprule Nodes&Scoped reuse&Value oracle&Stale unchecked&Stale released\\\midrule
65&473&475&745&0\\257&555&556&664&0\\
1,025&587&587&633&0\\4,097&599&599&621&0\\\bottomrule\end{tabular}\end{table}

The scoped rule released 2,214 requests. A full-value oracle could release 2,217. The difference illustrates the cost of conservative invalidation, including changes below the numerical threshold. The unchecked rule would have accepted 2,663 stale reads. Those counts depend on the explicit half-local, half-uniform request mix. A real controller that always reads the hand will get substantially less reuse. A readout that pools every graph node will almost always be invalidated.

\begin{table}[!t]\centering\caption{Actual object-track replay, median update time in microseconds. Rows are grouped by the observed number of tracked objects.}\label{tab:small}
\footnotesize\begin{tabular}{@{}lrrrr@{}}\toprule Number of objects&2&4&6&8\\\midrule
Full pass&33.7&35.4&37.3&40.0\\
Incremental&53.5&56.2&60.3&63.3\\\bottomrule\end{tabular}\end{table}

On object tracks the performance order reversed at every tested size. Across 3,600 timed samples the full-pass median and p95 were 37.1 and 47.6 microseconds, against 58.5 and 79.3 microseconds for incremental execution. The maximum absolute difference across all 3,840 updates was $1.11\times10^{-16}$. Bookkeeping for a few nodes costs more than a contiguous matrix operation. This result is important for table-top manipulation: small object-only graphs are common even if the surrounding map contains thousands of spatial elements.

\subsection{A fitted consumer on held-out object graphs}
We repeated the release test with a trained model to check that the result survives task-dependent weights and a nontrivial consumer. For each object we used its normalized position, planar differences over lags of one, three and six recorded frames, their norms and a constant. The network used the same three mean-aggregation layers with learned self and neighbor matrices, followed by a shared two-coordinate linear head. Its target was the object's displacement four recorded frames into the future. We took 96 uniformly spaced eligible anchors per episode. The first ten nominal episodes of each of the six tasks provided 5,760 fitting windows; the next five per task provided 2,880 held-out windows. Training used one initialization and 35 passes through the fitting set. Each AdamW update used 96 windows with a step size of 0.003. The exported weights and all held-out errors accompany the code.

Across 30 held-out episodes the fitted readout gave a mean endpoint error of 2.22 pixels, versus 2.60 for a zero-motion predictor, with equal weight per episode. The accuracy result establishes a modest forecasting use for the consumer. The runtime claim concerns its equality across snapshots. With uniformly sampled single-object reads and the same three-update delay, 358 of 2,790 requests passed validation. None differed from a fresh read by more than $10^{-10}$ pixels, whereas unconditional release would have accepted 2,336 changed readouts. The largest discrepancy between local and full recomputation anywhere in this audit was $1.78\times10^{-14}$ pixels. A stamp can reject an unchanged read through a dependent feature edit, which explains why 358 releases fall below the 454 numerically unchanged readouts. These expert traces carry no executed action from our decoder.

\subsection{The effect of a burst}
We independently perturbed $k$ uniformly chosen cell features per commit on a 1,025-node graph with fixed adjacency. Each of six values of $k$ used 100 measured updates after four warm-ups. At $k=16$ the incremental median was 239 microseconds and the full median 682. At $k=64$ these figures were 549 and 695. At $k=256$, affected rows covered a median of 90.7\% of all layer rows; local execution took 896 microseconds against 700 for a full pass. Editing all 1,024 cells increased the incremental median to 998 microseconds, versus 699 for full computation. These values bound the observed crossover between the sampled edit counts on this implementation. They do not determine a hardware-independent switching threshold.

\section{Queued Execution and Directed Dependencies}
A caller sees the receipt of a command, whereas the timings in Table~\ref{tab:tim} stop at the graph. We implemented a second path with a producer, one commit worker and a receiving thread. The producer timestamps a replay frame before copying seven consecutive positions into a bounded queue. The worker builds the motion features and directed incoming-neighbor table, refreshes the encoder, validates a read issued one frame earlier and evaluates the two-coordinate motion head if that read expired. The receiver copies both output coordinates before recording its timestamp. The next frame enters after this receipt. This single-outstanding-frame design measures software handoff at low load. Image acquisition, physical tracking, an actuator and overload queuing lie outside the measured interval.

The full comparator preallocates its matrix workspaces and uses dense adjacency multiplication for the mean aggregation. The local implementation follows the reverse of the incoming-neighbor relation to find consumers. For small graphs it scans the adjacency table at each layer. Both implementations read the same held-out tracks and network parameters. Each of 30 tracks contributed 64 frames in two traversals per mode, with the mode order reversed on the second traversal. We excluded the first four frames of each traversal from timing and compared the two-coordinate output on every frame.

\begin{figure}[!t]\centering\includegraphics[width=\columnwidth]{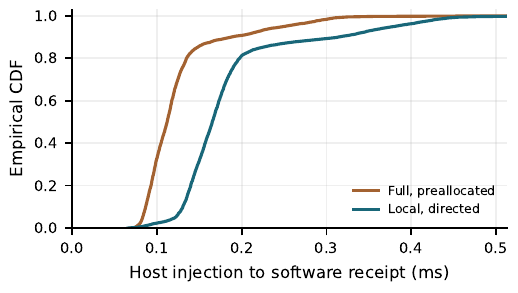}
\caption{Host latency over 3,600 timed messages per implementation. The plot shows the distribution up to just above the larger 99th percentile. The CPU includes the graph calculation, the motion decoder and both thread handoffs.}\label{fig:queue}\end{figure}
The full path took 0.112, 0.250 and 0.309 ms at the 50th, 95th and 99th percentiles. For local refresh the corresponding values were 0.165, 0.381 and 0.446 ms (Fig.~\ref{fig:queue}). The maximum observed output difference was $1.07\times10^{-14}$ pixels. Full recomputation thus retained its advantage for few-object graphs when the timer extended through the learned output and software receipt. Percentiles describe this host under the specified workload. A periodic camera and an actuator impose different queueing conditions.

\subsection{Ordering, replacement and undeclared state}
Two additional producer threads submitted edits to disjoint node sets, each with its own increasing source sequence. The commit owner interleaved fixed-size batches in reproducible round-robin order. It discarded duplicate or out-of-order sequence numbers, assigned a global commit version, and checked node generation, weight epoch and an auxiliary input epoch at release. Replacing a stable slot allocates a new generation. A weight change triggers a fresh full pass. Changing the declared auxiliary input invalidates outstanding reads even when the graph is unchanged. These rules serialize commits while allowing concurrent producers. Overlapping write ownership would require an additional conflict rule.

We exercised two directed graphs, each with 330 accepted commits, 58 dropped arrivals, 66 directed rewirings, 14 slot replacements, eight auxiliary changes and four weight changes. The independent oracle recomputed all layers before evaluating each delayed output. With 19 nodes and three incoming neighbors per node, one of 327 old outputs was released; unchecked reuse would have accepted 326 altered outputs. With 61 nodes and one incoming neighbor, 220 of 327 were released and unchecked reuse would have accepted 107 altered outputs. No altered output passed the guard in either run at $10^{-10}$ tolerance. The small acceptance count in the denser graph shows how rapidly overlapping dependency sets remove opportunities for reuse.

\subsection{Device placement and measured board energy}
A resident GPU encoder provides a stronger full-recomputation baseline than launching one GPU kernel from Python for each operation. We captured the dense three-layer forward pass and learned motion head in a CUDA graph for each object count. Its weights and workspaces remained resident. Each timed call started with host-resident float32 features and normalized adjacency and returned the motion output to the host. A CPU implementation used preallocated float32 workspaces and the same reductions. This boundary excludes feature extraction, graph construction and thread handoff. It includes both host--device transfers on the GPU. Across 30 held-out tracks, 48 frames and two passes (2,880 observations per device), the maximum output difference was $1.53\times10^{-5}$ pixels. At these small node counts the CPU median and 99th percentile were 20.3 and 36.6 microseconds. On the RTX 5060 Laptop GPU they were 107.1 and 259.4 microseconds.

We read the discrete GPU's cumulative energy counter before and after each of three 3-s blocks per condition, rotating the test order. Under a continuous stream of GPU calls the board averaged 25.41 W (25.12--25.82 W over individual blocks). The corresponding whole-board averages were 11.85 W during idle blocks and 10.08 W during CPU execution. Idle power varied from 9.64 to 15.51 W after preceding activity. These readings include background board activity. The counter covers neither CPU energy nor the robot's power supply, so the device comparison establishes latency and board draw without yielding a whole-system energy ranking.

\section{Which Dependencies Prevent an Incorrect Release?}
\subsection{A numeric value and the object to which it belongs}
A readout can be numerically unchanged and still refer to the wrong object. Suppose a tracker deletes a cup and later assigns the same array slot to a second cup with the same measured pose and embedding. The graph computation can legitimately return the same coordinates. A pending request to act on the first cup still carries the old identity. An equality oracle that compares two floating-point vectors alone would incorrectly count this reuse as valid. Conversely, a row stamp can be unchanged during a policy or coordinate-frame update even though applying the old output would use a different decoder or reference. These cases require different state in a request token.

For a one-row request we store $(i,g_i,s,e,c)$: the slot, its generation, the issue commit, the model epoch, and the epoch of a declared decoder input. A current token may be released when
\begin{equation}
 g_i^{\rm now}=g_i,\quad \sigma_i\leq s,\quad
 e^{\rm now}=e,\quad c^{\rm now}=c .
 \label{eq:contractr3}
\end{equation}
Separate model and context epochs allow implementations to re-evaluate all graph embeddings after a weight update while retaining a precise meaning for a local graph-row stamp. The weight epoch invalidates outstanding requests even if every row happened to evaluate to its former value. A context update may leave the encoder entirely untouched. Treating context as part of the decoder explains why a graph-level cache alone cannot certify its output.

Consider three successive commits. At issue, slot 7 stores an object at generation 4 and the decoder reads that slot alone. An unrelated remote row changes at the next commit. Proposition~1 allows the old value to survive if that edit cannot reach slot 7. At the following commit the tracker recycles slot 7 without changing its feature vector. Now the numeric comparison can still pass while the generation comparison fails. At the third commit an instruction embedding changes for every request. The context comparison fails even for slots that the tracker has never recycled. Checking only whether the last graph update touched slot 7 cannot recover either fact. The example also describes a read for which a change anywhere in the graph would be too broad a rejection rule at the first commit.

\subsection{Factorial release test}
We varied the number of incoming neighbors and the number of graph commits between issue and release while keeping the learned encoder and decoder from the object-track experiment. Each graph contains 64 slots and either one or three incoming neighbors per receiver. Eight initialization seeds per degree each generate 256 commits. Every commit issues a request at the preceding version, draws one event, and applies it. Event probabilities are 0.66 for a feature edit, 0.17 for replacement of an incoming list, 0.08 for an identity-only slot replacement, 0.055 for a decoder-context update, and 0.035 for a model-weight change. Rates describe this controlled stress test. They are not estimates of the frequencies on LIBERO or on a physical robot.

A feature edit replaces one row by a new random vector. Rewiring redraws the incoming neighbors of one receiver without replacement. An identity event increments one slot's generation and deliberately preserves its numeric features. On a weight event we perturb one scalar and recompute the encoder under the updated weights. On a context event the synthetic decoder offset changes by 0.013 in each output coordinate, leaving graph embeddings unchanged. The full model is evaluated from scratch at every commit as an independent numerical reference for the cache. To label a delayed request we compare its saved generation with the current one and its saved output with the current decoder output, using an absolute coordinate tolerance of $10^{-10}$. The reference therefore detects both a misidentified slot and a changed numerical output.

Requests are tested after one, four, eight, and sixteen commits. With $d$ commits of delay, a 256-commit run yields $257-d$ releases. Every policy sees the exact same delay, issue state, edit history, and requested node for a given trial. We evaluate six variants: reject any request with a version mismatch; check only the affected-row stamp; add the model epoch; also add the context epoch; use all four conditions in \eqref{eq:contractr3}; and release every request. The first variant rejects every deliberately delayed request. The last provides the count of stale outcomes when no condition is checked. The intervening comparisons give the incremental protection of each declared dependency for this event schedule.

\begin{figure*}[!t]\centering\includegraphics[width=.94\textwidth]{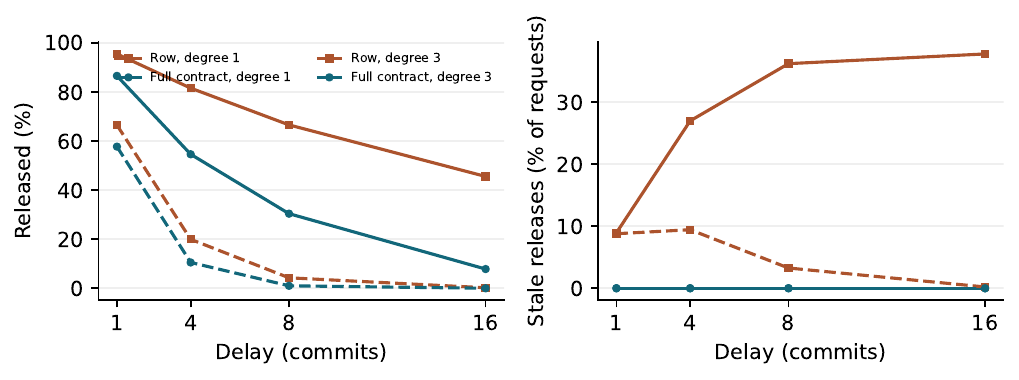}
\caption{Release decisions on directed 64-slot graphs. The denominator in both panels is the number of delayed requests at each delay and fan-in. The row-only rule ignores model, context, and identity changes. The full contract checks all three in addition to local graph edits. The two full-contract curves on the right coincide at zero.}\label{fig:guardr3}\end{figure*}

\begin{table}[!t]\centering\caption{Ablations at four commits of delay. Each degree has 2,024 requests. Each cell gives accepted requests / stale accepted requests. A stale result can have incorrect identity or a changed numeric output.}\label{tab:ablate}
\footnotesize\begin{tabular}{@{}lrr@{}}\toprule
Release condition & One incoming & Three incoming\\\midrule
Same global version & 0 / 0 & 0 / 0\\
Affected-row stamp & 1,652 / 546 & 404 / 191\\
Add weight epoch & 1,383 / 277 & 319 / 106\\
Also add context epoch & 1,113 / 7 & 214 / 1\\
Also add generation & 1,106 / 0 & 213 / 0\\
Unchecked & 2,024 / 917 & 2,024 / 1,811\\\bottomrule
\end{tabular}\end{table}

\subsection{Release yield, wrong releases, and conservative rejection}
For a specific delay, three denominators answer different questions. The release yield is $A/N$, the accepted count divided by requests. The fraction of all requests incorrectly released is $W/N$, where $W$ counts stale acceptances. The conditional fraction $W/A$ answers a different operational question and is undefined if the guard accepts nothing. We additionally count valid requests rejected: if the independent oracle calls $V$ requests valid, this count is $V-(A-W)$. A guard that rejects everything has zero stale releases and sacrifices all $V$ usable outputs. Reporting yield alongside incorrect releases keeps that strategy visible.

Table~\ref{tab:ablate} shows the four-commit comparison. Under fan-in one, 546 of the 1,652 releases allowed solely by the row stamp were stale. Adding weight and context epochs reduced the count to seven. All seven remaining incorrect releases disappeared when slot generation was checked. With all dependencies checked, 1,106 requests were accepted, compared with 1,107 that the oracle judged valid. One structural rejection was conservative under the chosen numerical tolerance. The analogous full-contract test at fan-in three released all 213 oracle-valid requests. These are paired re-evaluations of two streams. Adding the six policy denominators would count identical requests repeatedly.

Increasing the delay can make a correct guard operationally unhelpful. At fan-in one the complete check released 1,774 of 2,048 requests after one commit and 151 of 1,928 after sixteen. At fan-in three these numbers were 1,183 and zero, respectively. All 1,928 long-delay requests in the latter condition had genuinely changed according to the oracle. A more selective certificate could reclaim the few conservative rejections in other conditions; here no valid outputs remained to recover. The observed vanishing yield therefore has a different engineering remedy: shorten the issue-to-release delay, move the request closer to the current observation, or change the consumer's declared read set if its task permits. Faster integer comparisons by themselves cannot make an old output valid.

Across both degrees and all delays, the complete check released 5,053 of 15,984 requests with zero observed stale releases. The same commit history contributes to several delays, and successive reads within a trace share graph state. A binomial interval treating all requests as independent would give misleading precision. The proof supplies the logical basis under the declared model. The stress test exercises its implementation, including cases engineered to defeat a purely numeric comparison. The synthetic context and identity events test the semantics of the interfaces; they do not establish tracking accuracy or the effect of a stale action on a physical task.

\section{Where the Host Pipeline Spends Its Time}
The queued replay records four intervals with a common monotonic host clock. From injection until the worker takes the packet, input time includes a history copy and the first thread handoff. The graph interval includes feature construction, obtaining the current adjacency, and either local refresh or preallocated dense evaluation. The next interval includes the read check, selection of the returned command, and preparation of the next request. Sink time runs from that point until the software receiver records receipt of the two coordinates. These intervals partition the measured path without overlap. None starts at camera exposure; none ends at an actuator acknowledgment.

We recomputed the sum of the four intervals for each of the 7,200 measured packets and compared it with the saved end-to-end duration. They agreed to the resolution of the archived timestamps. Table~\ref{tab:stages} gives component means on all packets and on the slowest one percent of each mode. Selecting packets by total latency before averaging the components ensures that every term in a tail row refers to the same packets. The sum of independently estimated stage percentiles would generally correspond to no observed end-to-end packet.

\begin{table}[!t]\centering\caption{Host interval means in microseconds. Tail rows contain the 36 packets at or above each mode's 99th-percentile end-to-end latency. Component rounding can affect the displayed sum.}\label{tab:stages}
\footnotesize\begin{tabular}{@{}lrrrrr@{}}\toprule
Mode and subset & Input & Graph & Readout & Sink & Total\\\midrule
Full, all & 14.3 & 87.1 & 7.7 & 16.1 & 125.2\\
Full, tail & 53.9 & 244.1 & 17.8 & 44.1 & 359.9\\
Local, all & 15.2 & 148.2 & 4.9 & 18.4 & 186.8\\
Local, tail & 39.5 & 407.4 & 10.9 & 55.5 & 513.3\\\bottomrule
\end{tabular}\end{table}

\begin{figure*}[!t]\centering\includegraphics[width=.94\textwidth]{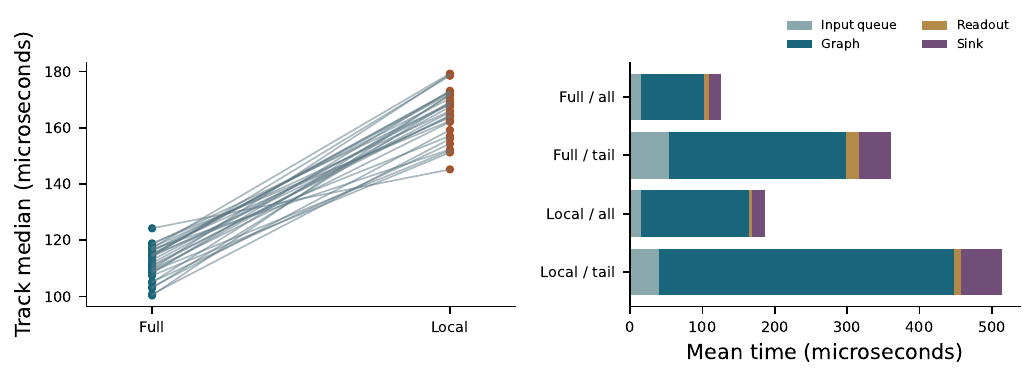}
\caption{Host measurements on held-out object tracks. Left, each connecting line follows the same track under both implementations. Right, additive component means from the same packets are shown for the full sample and the slow tail. The modes can select different packets into their tails.}\label{fig:stagegraph}\end{figure*}

In the full mode, mean graph time was 87.1 microseconds, against 148.2 microseconds for local maintenance. The readout interval favored the local mode by about 2.8 microseconds, much less than the additional graph work. Across 30 held-out object tracks the median total per track ranged from 100.4 to 124.3 microseconds for full computation and from 145.2 to 179.3 for local maintenance. The ordering therefore survives aggregation at the track level. This comparison uses the same machine, worker design and readout, and does not isolate one NumPy instruction or one memory access. In particular, refreshing a few rows still pays for finding graph changes and constructing receiver masks on a very small graph.

There is a deployment consequence to where the timer is placed. Moving the model to the GPU could shorten an encoder kernel while increasing host-to-device transport time; removing a thread handoff would address a different interval. On the observed host the full-mode slow tail spent an average 53.9 microseconds before graph computation and 44.1 after the readout stage. Even an instantaneous encoder would leave that time in the software path. The device benchmark starts from a prebuilt host graph, while the queued benchmark rebuilds it from observed positions. Their totals describe different boundaries, and summing those totals would count some computation twice.

\section{When Two Sensors Write the Same Slot}
The two-producer queue test assigns disjoint slots to writers. That arrangement makes serialization straightforward but leaves a separate question unanswered. If two sensors both report the position of slot $i$, which delivered observation defines the committed value? A cache can be perfectly consistent with a graph that was assembled in the wrong temporal order. We therefore test the selection of an observation before passing the winning value to the encoder.

\subsection{An explicit ordering rule}
Consider an absolute feature vector at capture tick $t$ with source identifier $b$. Within a tick each source writes at most once. For each slot we retain the observation with the greatest lexicographic key $(t,b)$. The source ID breaks equal-tick ties using a fixed priority. Duplicate delivery of the same message is idempotent. When ticks are comparable across sensors, the key defines a total order for these writes, and the committed value after a sequence of arrivals is independent of the arrival permutation. The claim follows by induction: the initial register is empty; comparing a new key with the current maximum keeps exactly the maximum of the old set plus the new observation. An event whose key loses the comparison produces no graph edit. A winning event with a changed feature uses the original propagation rule.

This design makes a particular modeling choice. Each message is a full observation of the same field and both sensors use a shared timebase. For two successive increments to a joint angle, retaining the last increment would discard motion. For mutually informative camera measurements, a fused estimator might combine both instead of selecting one. In those applications the committed state must encode the estimator's update semantics. A source ID only resolves ties deterministically. Its priority does not represent a claim about sensor quality.

We generated 16 schedules, each with two sources observing four shared slots of a 24-node directed graph. Each source captures one observation per tick for 96 ticks. Deliveries are delayed by an independently sampled integer from zero to nine ticks; selected packets are copied and delivered again one to five ticks later. The base graph has two incoming neighbors per node. At every arrival a reference enumerates the observations delivered so far for the addressed slot, selects the maximum true capture-time key, updates a reference graph, and recomputes the learned readout. The reference has no access to future deliveries. Thus a delayed but still in-flight observation cannot make an implementation fail this comparison. A system that also needs a maximum age relative to the exposure time needs an additional deadline check.

\begin{table}[!t]\centering\caption{Shared-slot arbitration on the same 3,360 packet deliveries. Wrong winner counts a different observation in the addressed slot. Output mismatch compares the readout at that slot on the complete committed graph.}\label{tab:sharedclock}
\footnotesize\begin{tabular}{@{}lrrr@{}}\toprule
Arbitration & Writes & Wrong winner & Output mismatch\\\midrule
Last arrival & 3,360 & 1,541 & 2,577\\
Per-source sequence & 2,256 & 783 & 1,585\\
Synchronized clocks & 1,732 & 0 & 0\\
One clock $+3$ ticks & 1,598 & 644 & 1,319\\\bottomrule
\end{tabular}\end{table}

\subsection{Order at the register and order at the readout}
The first comparator simply overwrites a slot at each arrival. The second discards any message whose sequence is older than or equal to the most recently accepted sequence from that particular source for that slot. The latter filters duplicate deliveries and reversals within one stream, yet a late report from source 0 can still displace a newer report from source 1. The third comparator uses the synchronized key. A fourth applies the same key logic after adding three ticks to source 1's capture clock, leaving the reference clock and packet deliveries fixed. No policy in this experiment estimates clock offset from the data.

As Table~\ref{tab:sharedclock} shows, source-local sequence checks reduced wrong ownership from 1,541 to 783 arrivals. They still left 1,585 numerical readout mismatches. A numerical mismatch can occur even when the addressed slot has the correct observation: a neighbor read by the network may retain a wrong earlier winner. Measuring the register alone would miss that propagation. Conversely a changed register need not alter a readout beyond the numerical tolerance. Reporting both quantities makes them distinguishable. Under the synchronized ordering, 1,732 winning writes sufficed and full recomputation agreed at every one of the 3,360 delivery boundaries. This result exercises the specified notion of the latest delivered observation. It says nothing about whether the last observation from either instrument is physically accurate.

The clock-bias test shows where the rule can fail despite a correct implementation. With a three-tick offset, 644 addressed slots had the wrong true-time winner and 1,319 readouts disagreed. Reprocessing the same biased timestamps in a different order would converge to the same wrong answer. If clock offsets are only known within intervals, two observations can be ordered confidently in physical time when their possible capture-time intervals are disjoint. Overlapping intervals require an explicit tie, fusion, or request-for-resynchronization policy. The experiments give exact evidence for the synchronized and deliberately biased cases; they do not calibrate clock uncertainty on a robot. A physical integration needs to log both capture and receipt clocks before numerical agreement at software release can be interpreted as temporal agreement at the actuator.

\section{Deployment Boundaries}
A graph-level release gate is relevant only when it covers the state read by the actual controller. A grasp policy might read an object embedding plus a wrist image, force signal, current instruction and robot clock. The graph certificate covers only the first item. The other inputs need their own version semantics or a fresh validation at release. If the policy consumes globally normalized features, all dependent rows must be included. An undeclared edge feature or a mutable auxiliary buffer invalidates the proof.

Applications in few-object manipulation can use a full pass while still maintaining read consistency by comparing newly computed read rows. In larger maps the sparse strategy is attractive when edits remain localized and queries often lie outside the affected neighborhood. The burst study motivates a switch to full recomputation when many cells change. Timing that switch on large maps remains necessary. A single owner commits writes and keeps all intermediate layers in memory. The generation test covers slot replacement in a fixed-capacity graph. For shared-slot writers, a capture-time ordering resolved conflicting absolute observations under synchronized clocks. Timebase alignment on a robot, fusion of simultaneous measurements and resizing the slot array remain open engineering issues.

LIBERO, RLBench and CALVIN offer environments for testing action execution \cite{libero,rlbench,calvin}. BridgeData V2 and studies of offline robot demonstrations illustrate the gap between logged motion and closed-loop evaluation \cite{bridge,robomimic}. A physical deployment would add timestamps at sensor exposure and actuator acknowledgment around the software interval measured here, and would declare the read sets of the deployed policy. The host measurements supply a baseline for that comparison.

\section{Conclusion}
A scoped stamp can answer a release-time question that a global version counter cannot: whether the declared graph input to an outstanding request has changed. Exact refresh and release checks held on both recorded object tracks and a recorded-motion map workload. For a few tracked objects, recomputing the full graph was consistently cheaper. For sparse edits on a large spatial map, the same local rule avoided most forward computation. The useful regime depends jointly on graph size, edit density and what the decoder actually reads.

\section*{Code and Data Availability}
The \href{https://github.com/LudwigWangMu/scene-graph-readout-reuse/tree/a5f7083fd58dac90146ce88c574d5ea48dd8eb9b}{\nolinkurl{scene-graph-readout-reuse}} repository contains the readout weights, packet records, board-energy measurements, and source-data identifiers. From this directory, \texttt{python code/prepare\_trace.py} retrieves the LIBERO inputs and \texttt{python reproduce.py} reruns the CPU workload. The README gives the separate GPU measurement command. From the repository root, \texttt{python verify.py} checks the published release and ordering counts. Latency and energy refer to the recorded host. The archived version is \href{https://github.com/LudwigWangMu/scene-graph-readout-reuse/releases/tag/v1.0.1}{release v1.0.1}.

\section*{Acknowledgment}

\IEEEtriggeratref{25}
\bibliographystyle{IEEEtran}
\bibliography{references}
\end{document}